# A challenge in A(G)I, cybernetics revived in the Ouroboros Model as one algorithm for all thinking

Knud Thomsen

Paul Scherrer Institute, Villigen and Würenlingen, Switzerland

E-mail: knud.thomsen@psi.ch.

**Abstract:** A topical challenge for algorithms in general and for automatic image categorization and generation in particular is presented in the form of a drawing for AI to "understand". In a second vein, AI is challenged to produce something similar from verbal description. The aim of the paper is to highlight strengths and deficiencies of current Artificial Intelligence approaches while coarsely sketching a way forward. A general lack of encompassing symbol-embedding and (not only) -grounding in some bodily basis is made responsible for current deficiencies. A concomitant dearth of hierarchical organization of concepts follows suit. As a remedy for these shortcomings, it is proposed to take a wide step back and to newly incorporate aspects of cybernetics and analog control processes. It is claimed that a promising overarching perspective is provided by the Ouroboros Model with a valid and versatile algorithmic backbone for general cognition at all accessible levels of abstraction and capabilities. Reality, rules, truth, and Free Will are all useful abstractions according to the Ouroboros Model. Logic deduction as well as intuitive guesses are claimed as produced on the basis of one compartmentalized memory for schemata and a pattern-matching, *i.e.*, monitoring process termed consumption analysis. The latter directs attention on short (attention proper) and also on long times scales (emotional biases). In this cybernetic approach, discrepancies between expectations and actual activations (e.g., sensory precepts) drive the general process of cognition and at the same time steer the storage of new and adapted memory entries. Dedicated structures in the human brain work in concert according to this scheme.

## 1. Introduction

Algorithms are everywhere, and (almost) everyone becomes increasingly aware of that now. Despite breathtaking recent advances in the demonstrated performance of Artificial Intelligence (AI), in particular of Large Language Models (LLMs), there are many prominent voices pointing out undeniable fundamental shortcomings of even the most powerful current approaches and programs [1–3]. On the entry page to ChatGPT, for example, there is a disclaimer acknowledging that ChatGPT "may occasionally generate incorrect information, may occasionally produce harmful instructions or biased content, and possesses limited knowledge of world and events after 2021" [4]. These statements could very easily also pass for a description of a human interlocutor; nothing special if many humans would not often dream of machine intelligence as always correct, unbiased, and sharp to name just a few positive attributes.

**2. A Challenge to AI**

In Figure 1, the challenge is to correctly classify that sketchy drawing as a whole, one "piece of art", and its interpretation comprising all aspects together. In a second, complementary, demand, an example AI is challenged to generate a similar image from a description. Some results of "snapshot-experiments" at different times are presented in Appendix A at the end of the paper.

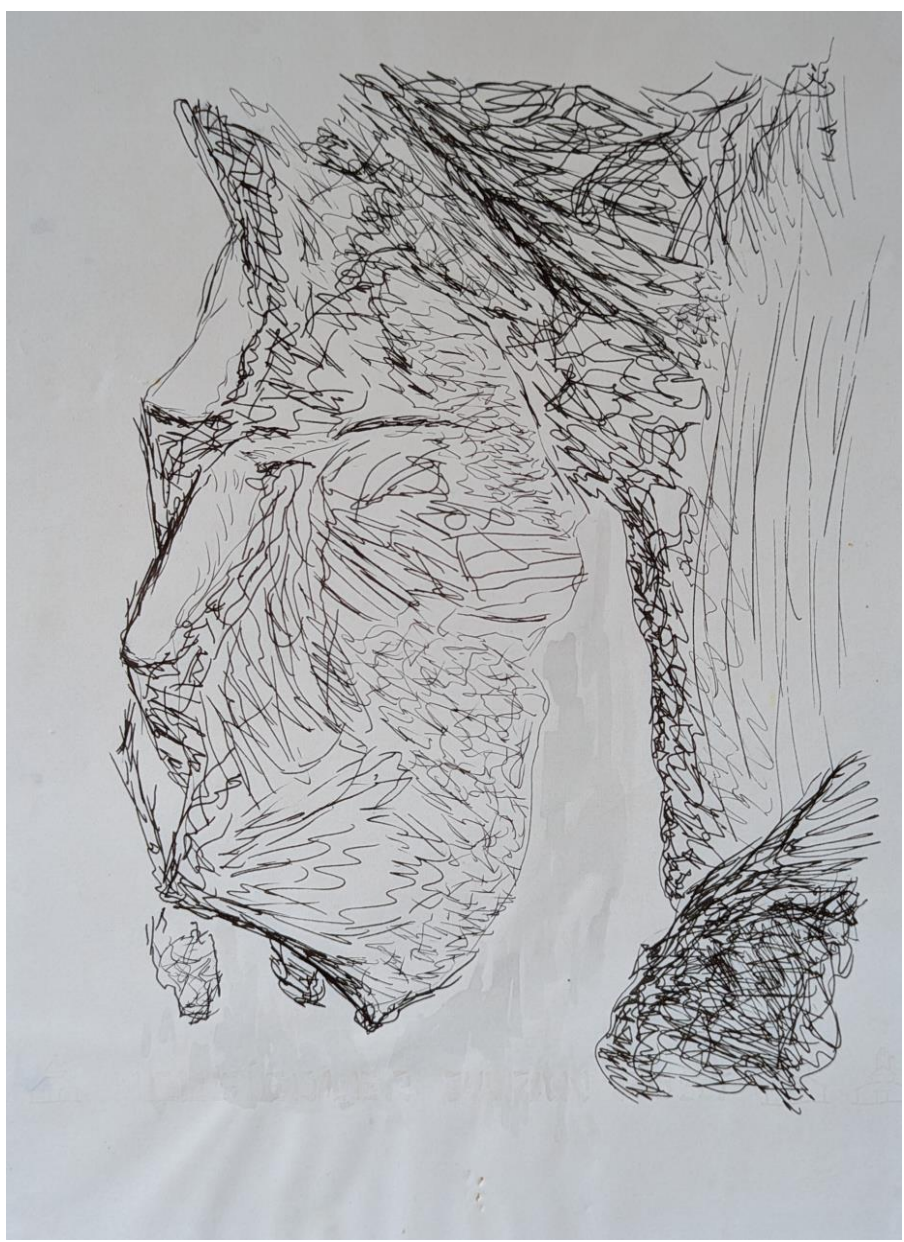

**Figure 1.** The original title of the drawing is "trollet og sitt hjem", in English: 'the troll and his home'.

## 3. Proposal for a more comprehensive approach

Given undisputed shortcomings and deficiencies of current AI as widely recognized and also cursorily documented in the Appendix A, a potential alternative approach is presented under the name of Ouroboros Model. At some point in time, *i.e.*, at some level of sophistication and autonomy (demanding some self-awareness), AI agents will want to have their voices heard; very brief conversations with ChatGPT are presented in Appendix B. In a somehow self-explicatory manner, it is attempted in the following to develop the arguments in an iterative and self-reflective way. This meta-perspective is further explained in Appendix C. Evidence for sketching an outline is drawn from a very wide range. It is self-consistently argued that in in order to follow an efficient way to progress this simply has to come as a first step before delving into any detailed scrutiny.

*The Ouroboros Model*

The Ouroboros Model has been proposed as a general blueprint for cognition [5,10]. It features only two basic ingredients: a memory structed in a (non-strict) hierarchy and a process called consumption analysis. The working of that underlying fundamental algorithm can be understood as a version of proportional control in disguise.

In a tiny nutshell: at one point in time, with a set of schemata available right then, an agent matches (sensory) input to these schemata and the one, which is fitting best, is selected. Comparing material and mold, there will most likely remain some discrepancies like features not assigned or subsumed/consumed as well as slots staying empty. On a short time-scale, attention will be directed towards exactly those attributes with the aim of improving the overall fit; – this is nothing else than a control loop geared at minimizing discrepancies between any current input and expectations based on earlier established knowledge. It has been claimed that with the right interpretation of existing schemata and incoming new sensory percepts, consumption analysis can be understood as an approximate implementation of Bayesian belief update [5].

Bayesian accounts in this context can explain more nuances than immediately evident. What does look like a simple conjunction fallacy, might turn out to be quite reasonable and rational when taking all circumstances into proper consideration [6]. Similarly, temporal discounting can be a wise attitude when normally dealing with a fundamentally uncertain and insecure world.

Connections to evolution can be drawn at minimum two levels; on a fundamental one, the genesis of creatures and their adaptations and capabilities follows a very similar path of emergence on demand with selection according to fit and usefulness [7]. Working out a formal mapping between evolution and Bayesian reasoning offers another related vein, and in "niche-construction", these two seem to merge seamlessly [8]. As Immanuel Kant already knew, pre-established concepts and constraints have a huge impact on what one can perceive, understand, and do, – and the other way round, *i.e.*, something like: "the conditions of the possibilities to experience objects are at the same time the conditions for the possible objects of this experience" [9].

Some more details on the working of the Ouroboros Model have been presented in a series of papers [10–12]. Here, aspects demonstrating its correspondence to analog control are shortly highlighted. It is important to stress that only an outline can be presented while not diminishing the role of due digital or formal implementations of the architecture. On the contrary, this policy only bears witness to the fully self-consistent approach advocated: beginning with an overall (approximate) schema, highlighting slots, which are deemed relevant and (partly) empty or discrepant. Adaptations often will materialize during the process of iterative filling-in; – for static input, and even more so when there are significant variations over time. In cases where massive changes are necessary, new schemata will be created [11]. Additionally, repetitions and similarities will lead to the grinding-in (and the abstraction) of proven useful concepts, structures, and procedures. Some measure of regularity seems indispensable. In extension of the well-known anthropic principle beyond cosmology, no overwhelming chaos could ever be a cradle of cognition; rather, only behavior, which can be captured by rules, can actually develop and prevail [7]. Without a minimum of stability and repetition, neither life nor human observers with sophisticated mental structures could ever evolve.

Both, analog and digital, characteristics can neatly work together; this has been dealt with in a dedicated paper [12]. Dichotomies are seen as a first step towards organizing some apparent tangle into distinguishable parts. With growing differentiation and understanding of dependencies, finer nuances become discernable, rendering the original b/w dichotomy a crude approximation. Nice to observe, even in the (currently standard) implementations, when purportedly intelligent behavior is simulated on digital computers, countless weights of synapses in large artificial neural networks are carefully tuned during extensive training to establish finely grained and thus basically analog connections.

In the light of the Ouroboros Model, "art" can quite generally be characterized as having added something novel to standard schemata and expectations, plus, especially, also weakening some selected correspondence(s) with "a real natural thing" in some type of (re-) presentation, *i.e.*, intentionally discarding some features and dimensions from what would "naturally" belong to an "everyday" entity while staying consistent to some extent (even if fully realistic, a painting of a scene, e.g., is not the "prime reality" of the original scene itself). It follows that a certain level of intended discrepancy to known natural mental models is required for any artifact to count as eligible as a piece of art, and reality monitoring is in a special and (partly) suspended mode for full appreciation.

From this perspective, AI in its beginnings started as an artistic project. Not only in hindsight it is clear that important aspects were and still are missing. The 1956 Dartmouth Workshop, widely considered foundational for the field of AI, did not include any substantial contribution by Norbert Wiener, then certainly one of the most qualified experts on control, automata, communication in animals and machines [13]. Maybe, this omission or rejection was a wise decision at that particular point in time given the then prevailing constellation. This does not mean that it makes sense to continue neglecting a possibly major source of inspiration.

Now, it is certainly appropriate trying to rectify some historical neglect or animosity and look what cybernetics might have to contribute facing the present-day obstacles for achieving human level general intelligence, especially from the point of view of efficient algorithms, as is claimed here.

## 4. Cybernetics reloaded

Stripped to its bare essentials, cybernetics deals with control [14]. It starts with simple biological and analogical mechanisms to keep certain features, e.g., the concentrations of some nutrients, within acceptable bounds, and it reaches up to the steering of complex reactions taking into consideration many different attributes and distinct levels of organization. Complexity increases in particular when feedback pertaining to the controlling system itself is taken into account. Herman Haken's Synergetics describes sophisticated extensions to the basic layout and also Nicolai Hartmann's ontological theory [15,16]. Emergence does happen, and it can explain a lot. It has recently even been claimed that impressive emergent capabilities have (already) been exhibited by LLMs [17].

At any one level, consumption analysis highlights specific discrepancies between an existing schema and actual content, which are determined in a matching / monitoring process. The results are used on different time scales for direct further search or action, and for setting a longer lasting affective tone of a situation and entailing a suitable bias for an agent [10]. Seeing the first as analogue to the operation of a simple (linear) control algorithm is straightforward. The immediately following question then is, what about more intricate control systems like PID [18]? For autonomous systems, the necessity of anticipations has to be emphasized; it enables living (moving) beings to prepare answers proactively even before a predicted demand fully materializes. A detailed overall account goes beyond what can be delivered in this short sketch; this is posed as another challenge to AI (early anticipation surely is more relating to differentiation while bias to integration).

Following the Ouroboros Model, some form of equilibrium and adequacy (non-disturbing deviations from appropriate set-points) are the core goals at all levels from physiological needs, to bodily sensations, to abstract argumentations, e.g., in law suits, or the delivery of "permissible" weapons to a country under attack, to Free Will.

Adequacy itself is also an abstraction, and it is context dependent as epitomized in Carl Sagan's emblematic adage: "extraordinary claims require extraordinary evidence". This was evident to thinkers long before as expressed in the principle of Laplace: "The weight of the evidence should be proportioned to the strangeness of the facts" [19]. An appeal to "proportionality" is often issued in (political) discussions, and as some last resort when humans are confronted with atrocities and "just" measures of retaliation.

According to the Ouroboros Model, the potential mental processing power of an agent is ground-laid in the available knowledge, *i.e.*, the number, complexity, and elaboration of the concepts at her disposal. Schemata, their number of slots, the level of detail, the depth of hierarchies, degree of connection and interdependence of the building blocks, and the width, *i.e.*, the extent of some main schemata and their total coverage from the grounding level to the most abstract summits, determine what can be thought of efficiently; (quasi-)global adequacy, coherence, and consistency are crucial. Sheer performance at a single point in time arises as a result of the optimum interplay between these structured data and the effective execution of the described processing steps, in particular, self-referential consumption analysis.

With schemata as clearly distinct entities, "compartmentalization" stands in stark contrast to a rather indiscriminate associationism, which seemingly still lies at the core of the current big artificial neural networks. Compartmentalization provides the very basis for effective monitoring and meaningful error checking. Not only what is normally part of a given schema is specified but also what definitively (at that point in time) does not belong to it. Observed on a meta-level, the absence of an expected signal or feature is a valid feature in its own right.

Negation is a tricky concept/operation, for humans and even more so for current AI [20]. A "not-tag" is attached (as positive information) to a (major) constituent feature of a schema, and this allows for a lot of ambiguity. Recent improvements in chatbots in this respect can be traced to feedback from human instructors during training [21]. In general, in the real world, no unique "opposites" are well-defined. No straightforward nor very meaningful "tertium non datur" can therefore be expected in interesting contexts.

A knowledge cut-off, *i.e.*, a pretrained model knowing only of training material up to a certain point, severely limits the usefulness of LLMs and can lead to hallucinations. It comes as no surprise that cutting-edge attempts of improving on the performance of ChatGPT, in particular with respect to truthfulness, employ human feedback for reinforcement learning and using learned thresholds as a proxy for an "oracle" [21]. The reward model is trained with supervised learning and a relative ranking process for answers in a specific context by humans. Consumption analysis, sometimes harnessing human (corrective) input, intrinsically delivers something of that sort.

Like with any type of alphabet, versatile building blocks enable very efficient stepwise construction of almost infinite compositions, harnessing incremental and possibly nested procedures [22]. "Anchoring" at solidly verified specific points certainly is a good idea in principle; considerations can go astray when nothing of true relevance is available, and anchoring can then distort all kinds of (human and AI) actions. Overtly reporting external sources and associated argumentations certainly increases transparency and acceptability of and for any actor.

Quality-checked building blocks can constrain any construction and prevent going too far astray. They can provide some transparency as confirmed memory entries make it also possible to verify well described facts, like the details of the death of Otto Selz, e.g., by just looking that up in Wikipedia (quality-controlled by humans (which does not guarantee being 100 percent correct either); ChatGPT obviously did not do that, and got it wrong when asked about Otto Selz in February 2023).

Considering alternatives and, linked, push-pull processes make up a central part of an overall cybernetic conception. Beyond the most basic on/off controllers, there are many variants conceivable and necessary for adequate and fine control. Mismatches can be minor and negligible with almost no impact or change of action needed, or, they can be so fundamental that an ongoing activity has to be immediately terminated and some better alternative has to be found. These switching points are determined by active thresholds for living beings, some ingrained over eons of evolution in the bodily hardware (e.g., reflexes) while others are learned as results of prior occasions or observations during the course of growth or unfolding action, e.g., as potential turning points in a sequence of steps. There surely are some hard boundary conditions but determining appropriate thresholds itself often is a recursive process (second-order consumption analysis).

No matter at what level of abstraction, in the end a situation, a fit, will be evaluated as satisfactory or not satisfactory, with a wide range of intensity for a "feeling" of success or failure. Abstracted in a meta-perspective, this is seen as the basis for the fundamental concepts of "good" and "bad". This dichotomy is intrinsically linked to survival and evolution, thus truly foundationally imprinted and subsequently overshadowing in a sense everything and all the time, every percept or action of living beings including humans.

## 5. Common sense and understanding

What is meant in Figure 1 is immediately clear to humans −even before learning of the title − when looking at the picture in landscape and in portrait orientation: a drawing of a Norwegian landscape with a small fjord between steep mountains / an ugly face with a prominent nose. Humans do not see anything ambiguous in Figure 1; two distinct interpretations are fully valid when viewed separately, and there is a common one as implied by the title. (As an aside, it would be interesting to carefully test his presumption on a wider statistical basis with human subjects with different backgrounds.) Embedding and first-hand symbol grounding on a most basic bodily level are absent in the probed AI [1–3,22]. Some "sideways" as well as, similarly, "upwards" connections, *i.e.*, linking percepts to medium or highly abstract level concepts appear missing.

Humans often have some idea what an appropriate answer to a question could be, and, especially, what could be ruled out, even if they do not know any well-proven answer. This is a manifestation of common sense, accompanied by a gut-feeling, an intuition, how hard a problem is and how to solve it (or not). The global monitoring signal of consumption analysis yields values for the goodness of fit with experience (on all levels of sophistication) ranging from: fully accepted solution available to completely impossible, or: no idea. Even the latter constitutes valuable information in itself for subsequent activities, also in cases when it only tells to forget about trying and to safe energy. ChatGPT developers had to resort to human teachers to have their system learn to say "I do not know" [21]. Transparently declaring an impasse most often is much better than filling-in some superficially fitting content, as was obviously the case when first asking ChatGPT about Otto Selz. Common sense, according to the Ouroboros Model, in any case quickly yields and works first with something equivalent to a patchy sketch or draft ("Schematische Anticipation" in the words of Otto Selz [23,24]).

The Ouroboros Model, stiving for a most comprehensive picture, self-consistently relies on self-reflective iterative procedures for incremental self-steered growth. For a deliberate sketch, (initially) detailed matches cannot be expected and are neither demanded to start with. The combination of many different approaches and facets self-reflectively seems most promising. Deeper understanding and better explanations can be visualized as meaning a bigger "diameter" of a loop from bottom to top and back in the edifice of connected and interwoven schemata. Large loopholes in that web of concepts on the other hand side cannot be tolerated for any truly convincing account; in the picture of a net this would mean some small (enough) mesh size. In all circumstances, an acceptable explanation has to encompass an appropriate minimum diameter.

The elegance of a theory is determined by the clarity of the underlying assumptions, and it raises when only very few are required as its foundation. On the other side, if all arguments for an explanation rest on one basic element, highest caution is advised at the very least. Anything can be "explained" when some basic key concepts are tailor-made for that purpose. A falsifiable model with solid grounding, wide embedding, transparent structures, and causal mechanisms is much more difficult (and valuable). It goes without saying that a (meaningful) precision for a fit is of paramount importance; *i.e.*, the best (one-size-fits-all) explanation is no explanation when there are too many open ends.

Any conclusion, simple percept or resulting from sophisticated considerations, gains much credibility when there are multiple and independent paths leading to that result; the higher their diversity and when starting from different venture points, the better. This is an argument for beginning with a big variety of views for any discussion, a plea for plurality is the natural conclusion [25].

So, what about "understanding"? In the light of the Ouroboros Model, something is understood only if there is a complete model taking care of all (essential) aspects. In shades of gray, a certain minimum correspondence between "reality" and the "mental model" is demanded containing (the most important) (in-depth) details of structures and processes. In any case, mere replication is not sufficient. Not to offend logical rules, just the same as simple rewording, does not suffice. In chemistry for example this might mean different synthesis-paths to arrive at the same final compound.

As a trivial corollary, solipsism falls flat, at least in this common-sense perspective as sketched above. The same applies to a modern version, the "simulation hypothesis" [26].

Except for very well-regulated cases like formal logics in finite domains, a principal uncertainty of what really belongs to every question is inevitable, and no general eternally correct and valid answers can be expected.

Rules are abstractions, and their projection back to any specific single case is by no means guaranteed to be really meaningful.

Compartmentalization, which strongly limits the applicable content at a particular point in time, often enforces a trade-off.

Considering all possibly relevant features is the best one can aim for. In interesting cases, there most often is no external assurance whether something is sufficient for a predetermined level of fit or certainty. On the other hand, demanding a minimum of relevancy guards against the problem of "tacking by conjunction", which plagues simple orthodox Bayesian confirmation theory [27,28].

It goes without saying, hidden gaps in chains of reasoning should be avoided (but sometimes, we just do not know better). Openly admitting some lack of convincing supportive evidence certainly is better than confabulating hallucinations. There undoubtedly are (e.g., religious or artistic) contexts, where a leap of faith is unavoidable, even a fundamental requirement; this might not be to the taste of everyone. On the other side of the coin, tautologies cannot tell much new, and they generally will not be considered interesting.

Leaving some open ends, at least hinting at them, and allowing for a minimum of uncertainty, gives the necessary freedom for expansion or compromise, e.g., in case of disputable arguments. Noisy inputs and fuzzy borders of concepts will in this sense help easing transitions between related schemata (basins of attraction). Even if a threshold is not really exceeded a transition process akin to quantum-mechanical tunneling can be enabled by noise and some form of stochastic resonance.

If ready-made schemata are available, this allows for quick responses. Well-established schemata absorb input readily thus focusing attention; no iterations or lengthy considerations are required. On the other hand, lacking firm direct connections, links have to be iteratively searched for, constructed, tried, scrutinized, and verified. Strongly exaggerating these distinctions, diverse dual process models have been proposed [29].

Generally, full understanding demands a model, which covers all relevant levels of features and concepts (schemata). Understanding in turn is the basis for correct and exact anticipations. Anticipatory action asks for responses starting already at first signs of an event, leading to earlier and earlier onsets of reactions as familiarity rises.

Schemata are the very basis of every understanding; they are the organizational building blocks laid down in memory. In any case, for making a step (forward) some first foothold is required. It is claimed that, except maybe in deep meditation, some vague schematic preconception(s) will always be activated. Filling-in of slots and elaborations then are the subsequent steps when triggered by an external or internal event. The first activated schema might turn well out as not appropriately fitting. This can provoke minor updates or the establishment of adapted or completely new concepts [11].

Edifices of thought can break down, *i.e.*, when adding new information renders a first mental model obsolete and another interpretation much more likely. This effect is used inter alia in jokes [30].

Seen from a distance, inconsistencies with established prior knowledge (assumptions, guesses) propel and fuel any development and improvement. The Ouroboros Model thus sheds some light not only on the unavoidable occurrence of discrepancies but also on their necessity for growth and all interesting positive development. This, again, holds true at all levels, perfecting perceptions and movements, and also when dealing with some most abstract questions as, e.g., concerning dualism and Free Will.

However ground-laying and important first guesses are, they might turn out problematic or even wrong when more thoroughly scrutinized. A recently brought up example is the concept of "fairness". A plethora of meaningful and plausible definitions have been proposed [31]. When attempting to strictly formalize these it has been found that three appealing and innocent looking conditions cannot be met at the same time, except for rather trivial special cases; trade-offs are inevitable, regardless whether it is humans or AI to decide [32]. This does not really undermine that fairness can be seen as fundamental for justice [33,34]. It rather shows that no external God-given standard is available, and for humans, the applicability of whatever label or brand name can (and has to be) agreed upon in a specific context. At the extremes, highest precision and useful flexibility are mutually excluding each other.

An interesting example concerning the utility of heuristics has been given relating to exactly the very concept of heuristics; the fertility and huge impact of that conception can to a good part be traced to its imperfectness and the persistent lack of any precise narrow definition [35]. In the terminology of the Ouroboros Model, heuristics would correspond to sketches where only selected and most eye-catching features are taken into account.

Abstraction and sketching can anticipate a frame, which later turns out to be of little direct use, anticipations can lead astray. As an example, it simply would not tell anything (except a lack of engineering background) if a philosopher could think of an airplane built completely from lead; dreaming of zombies is the same (I maintain).

The Ouroboros Model confidently and proudly embraces functionalism, albeit not a trivial ("one-dimensional") version but one, which takes as many as possible (deemed important) dimensions, aspects, and constituent conditions into self-reflective consideration. Widest reaching consistency is the crucial criterion for learning and (considerate) action as described above and in several papers before [10–12,22].

It has been hypothesized in different proposals that all mental processes can be captured in sophisticated (production-)rules and relatively simple algorithms, which heavily rely on iteration and recursion [36]. The Ouroboros Model explains linear if → rules as abstractions from filling (remaining) slots in an otherwise well-defined schema (thus flexibly subsuming production rules while dramatically boosting efficiency in general).

## 6. Brains, natural and artificial, consciousness

Especially in cases when there are powerful constraints, e.g., a preconceived convictions lying unquestioned at the bottom, formalizing sometimes cannot help; there simply might be no solution possible within that given frame. An example could be John Searle, who, when discussing his famous Chinese room clings to "biological naturalism" and denies other than biological hardware to possess the "causal powers" that permit the human experience of consciousness [37,38]. John Searle in fact acts from an ideology, very comparably to what he purports of supporters of functionalism. No doubt, nobody would mistake the Chinese room for a human in a direct encounter; − just thinking, e.g., of the time it would take to receive any meaningful answer. Quality often arises from quantity. More is different; a single molecule of water is not wet [15,16]. "The whole is something beyond the parts" was already clear to Aristotle [39]. Other important and similarly decisive factors are speed and the mastering of (nested) contexts, in particular, negation.

In terms of a neural implementation of the Ouroboro Model, it is hypothesized that cortex (areas), hippocampus, and cerebellum are each specialized for specific tasks like memory or action (bodily and mental movement) [40–42].

Simple if → then relations, which do not require any sophisticated consideration (e.g., reflexes.. habits), will be relegated to automatisms in the basal ganglia allowing very quick / automatic reactions. Shortcuts will thus be implemented for often-used building blocks (e.g., movement schemata, like "assembler routines"). In bigger vertebrate brains, basal ganglia are primarily seen as "driving" and "power-" stages of/for higher level cortical (and cerebellar) areas steering effectors, controlling and regulating the processing in the diverse structures. Most importantly, they modulate (not only cortex areas), enforce gains and thresholds, e.g., relating to importance and speed. Synapses, which are marked for memory entry, will be strengthened, and an "emergency stop" ("veto") can effectively be realized. A neural implementation of a sophisticated effector-algorithm for fast control, *i.e.*, stopping, has just recently been described [43]. Push-pull strategies for fine control of movements in animals and humans appear to be employed ubiquitously [44,45].

A common misconception is that a "primitive brain" sits below an "advanced" cortex. This is like calling (steering) wheels old and primitive as they also existed before modern cars. Some basic functions need specific components, maybe with different details in diversely adapted implementations. "Modern", enlarged cortex volumes just add flexibility by making more options available, for perception, action, thinking, and self-reflection.

Learning, according to the Ouroboros Model, is based on fast ("snap-shot") and slow ("grinding-in") contributions [11]. Optimizations of connections do happen iteratively and often incrementally. The simple idea in this respect is that there are two possible ways to connect an input pattern with an output for tuning: one is backpropagation, and the other one is recurrency, *i.e.*, going the full circle a second time by reiterating the loop and processing that (or similar) activation again (quickly and after some time at a second related occasion) in the forward direction while taking into account all earlier results. Repeated runs can in particular harness the global feedback signal from consumption analysis and also distinct markings attached to specific components (slots, features, attributes,..). This can happen on a fast timescale reinforcing recently successfully employed synapses, and over longer timescales when positively tagged content is preferably integrated into long-time memory.

In addition to enabling efficient consumption analysis, clear and distinct "compartmentalized" records (schemata) allow for efficient indexed storage and retrieval but also require interpolation for meaningful use in real-world settings (most probably not only in vertebrates) [40,41,46].

It has been argued that any agent, who has to take some responsibility for her own functioning, e.g., caring for energy supply and avoiding errors and predators, at a certain level of sophistication mandatorily has to consider "household parameters" pertaining to the system itself. Subsequently, with the addition of some first basic intrinsic motivation to "survive" (like all animals obviously have even long before they master much language) any cognitive system will inevitably abstract/develop higher level aims and goals and a rudimentary form of self [47,48]. This awareness can be understood as the roots of (self-) consciousness. For this, details of implementation do not really matter, the most important ingredient is self-reflectivity. Higher order personality activation (HOPA), a form of higher order global state, of course, would be rather different for living beings and humans, communities and organizations, and, especially, for artificial agents. While robots might be somewhat closer to humans than pure software agents, the cognitive basis would be very similar. HOPA includes the highest-level goals and values of an individual for the person herself -- and for outside observers(!). Efforts to endow robots with self-awareness harnessing inner speech are underway [49].

John Searle is right, the detailed intentions certainly would not be identical for humans and artificial agents but in direct analogy they should be seen as equivalent with respect to their fundamental importance for any particular individual [37,38].

Generally, features are of different relevance and centrality for different schemata. Airplanes do not flap their wings like birds or butterflies; still, nobody doubts that they do fly. Like no "élan vital" is required to principally understand biochemistry and life, no fundamental difference is seen as to whether a living brain or silicon forms the hardware substrate for cognition and self-reflection (likely except the attributions by others).

LLMs are built as rather plain artificial neural networks, *i.e.*, statistical models, which predict what objects (words, in this case) usually follow others in a sequence. The achieved impressive performance can be attributed to the fact that human words are symbols for concepts, which often stand for rich contents and bear significant meanings for humans (as speakers and as listeners / actors and recipients). Recent transformer architectures effectively include some type of top-down influence. Nothing mysterious, human children regularly learn a language from their parents, and inner speech often is advantageously used by children (and adults) when performing difficult tasks.

Further adding to the recognized power of inner speech, its provision has been proposed recently to render the workings of robots easier to understand and trust. First tests show that this is indeed appreciated by humans, and reported inner speech influences the participant's perceptions of a robot's animacy and intelligence [49,50].

In ChatGPT, as in other current deep neural networks, obviously lacking explicit sophisticated hierarchical structure, verified building blocks are apparently missing [51]. For discriminating human users from machines, simple CAPTCHAs (still) can be used. It is tempting to see a parallel with humans switching to a similar mode of confabulations without the usual structure or constraints imposed on neural activations and connections by higher level schemata when consuming mind-altering drugs like LSD [52] (trivial corollary: CAPTCHAs become difficult for humans when intoxicated).

Humans as the examples of self-conscious beings closest to us personally, are embodied in a particular way. The private experience of qualia of an individual has been claimed as peculiar characteristic of humans (and other living beings).

According to the Ouroboros Model, qualia are abstractions, percepts of/for an individual and linked to her body, which cannot be other than private to that healthy(!) individual. Still, in exchange with others, similarities in perception and functioning (based amongst others on the common heritage from biological and cultural evolution) allow agents to agree on shared labels for individually experienced content.

No insurmountable difference to other self-monitoring and communicating agents is visible (except when postulated at the outset). Human societies have developed a great many diverse cultures, and yet, it is hard to imagine how these might be extended to fully embrace AI, artificial agents. Developing an attitude towards artificial agents as expressed in Ubuntu for humans among them might turn out difficult [53,54].

Most probably, individual agents acting in real-time in a dynamic world of whatever type need intermittent off-line phases for "housekeeping", *i.e.*, consolidation of useful stuff and discarding of inescapably accruing "data-garbage", especially during sleep [55]. Most interestingly, clever birds and octopuses not only sleep but also seem to dream similarly to mammals despite (apparently) rather different brain lay-outs [46,56,57].

## 7. Reality and truth, Free Will

The number of alternatives, which are available for understanding a certain state of affairs has been found to explain the convincing power of conditionals, counterfactual reasoning, negation; e.g. [58]. The question then might be, quite generally, where this leaves reality and truth. The upshot following the Ouroboros Model is that reality exists even when not observed, but details are to some extent in the eye of the beholder. The existence of an independent reality cannot independently be proved completely, *i.e.*, from a truly outside perspective. Laws of physics and whatever rules are abstracted from repeated successful applications, and they tend to live a life of their own, knitting and sometimes also cutting links to the special cases, which first allowed their distillation. According to the Ouroboros Model, relevant and real is something, which has an unquestionable effect (– for somebody in particular, but not always necessarily so); this, obviously, is also the basic stance of the ethics experts, who compiled a very timely assessment of the challenges imposed by AI to humans, e.g., writing about responsibility [59].

Most important here seems to be that humans necessarily grow up in some form of community and society. There they learn, e.g., roles in their culture by being taught and also by imitation, and they experience others and themselves as individuals. Explicit yes / no reinforcement feedback is delivered, especially for important topics. After quite some learning, humans ascribe and they are ascribed individual personality, subjectivity and also Free Will. A large impact of external attribution can be seen from its reversal; undermining the belief in Free Will made participants in a test feel more alienated from their true self, and it lowered their self-perception of authenticity [60].

The most materialistic science known, *i.e.*, physics, has taught us that reality appears only fully real in a rather limited sphere (at least with a clear (preferably causal) connection to observables) centered around values directly accessible to our senses. Experts and non-experts discuss what "real" could mean in the foundational quantum realm, e.g., for the concept of time; nothing is accepted as real without leaving some form of a trace, no clock is a clock without some memory [61].

Free Will is real. It is real in the sense that this abstracted conception exerts very tangible effects. It is amongst others foundational for an understanding of responsibility [59]. Directly linking Free Will to lowest level substance categories means committing an error of confounding and short-circuiting the appropriate very distinct levels [15,16].

"Free" commonly means not forced by foreign factors, it does not mean completely indetermined nor random.

My will is free when I have a chance (*i.e.*, sufficient resources and time) to consciously weigh alternatives and when I can choose one option in the end without being forced to that decision by obvious external circumstances or other compulsions. It is not some fundamental determinism or chance, which blindly rule, but it is me who decides, *i.e.*, I self-reflectively take into proper account my values and goals, my motivations, my intentions, my experience in my current situation, and so forth. My conscience and also my unconscious bodily and mental basis are mine, personally, and I only go through some lengthy deliberations, when I consider that demanded and the effort worthwhile.

A little bit of luck, additional (unexpected, also random) input can boost the freedom of a decision when making more advantageous options available.

(Overall) consistency is the aim and the measure; consumption analysis is amongst others an efficient way to implement a veto if I notice some important contradiction [47].

As there is no way to know all relevant factors in detail in advance (and most often probably not even after the fact), free decisions are never fully predictable, -- not for oneself before any thorough considerations and evaluations of options have been performed, and even less so for an outsider.

Non-predictability is not the same as randomness. Any seemingly random action does not mean at all that someone/something is "free"; non-predictability can result from deterministic processes involving some fundamental limit or statistical uncertainty deeply ingrained in a process.

There is no "absolute" freedom in the real world, and there cannot be; there are shades of gray, and scale / level of abstraction do matter. Despite all necessary grounding, higher levels of abstraction can and often do break free from (some of the) possibly tighter restrictions effective at lower levels [15,16].

It has been argued that as soon as self-steered and self-reflective growth is accessible for any agent, the predictability of her actions diminishes, and the actor thus also gains freedom in her deeds (and omissions) [25,34]. This applies to humans when they grow up and it will apply rather similarly to artificial autonomous agents. As with humans, the hope is that with careful responsible "upbringing" and education any agent capable of self-reflection, self-consciousness and some autonomy can be successfully directed to strive prudently for mutual and common benefit [25,34].

The hope then is that truly intelligent AI will pay heed to a negative imperative ("given an inescapably limited overall frame, violence has to be avoided as a result of reflected self-interest") as has been argued for prudent human beings.

# 8. Conclusion

Selected short experiments with clear results are collected in Appendix A. A general lack of symbol grounding and common sense, which for humans is "naturally" given by their embodiment, is identified as one reason for the non-convincing performance of even the most powerful current approaches to Artificial (General) Intelligence; comprehensive embedding and meaningfully considering highly abstract concepts also appears to be mostly missing. At the same time, present-day AI approaches do not seem to be flexible or creative enough to find two options or its full meaning if asked for an interpretation of the drawing in Figure 1. This goes hand in hand with the wide absence of versatile building blocks, organized in some sort of hierarchy of concepts and abstractions, and, claimed as a direct consequence, lack of "common sense" reasoning [22].

Here, it has been attempted to coarsely sketch that the Ouroboros Model could offer an overarching framework providing just that, which is now found to be still lacking in AI. Not a full-fledged "theory of everything" can be presented at this time but an outline of how all of the experienced and accessible reality can be perceived and in a sense thus functionally established, and how basic cognitive processes could be understood on the basis of one truly fundamental algorithm and its accompanying organizational (data-) structures; – a contribution on a meta level, for everyone, applicable for all agents with a prime focus on (self-)consistency, self-referral, and autocatalytic self-guided growth.

Including top-down guidance from the beginning, humans immediately grasp Figure 1 as a sketchy drawing. Setting the mind-frame to drawing and sketch, certain detailed features lose their weight. On the foundations inherited over eons of evolution, for humans, directions like up and down have their intrinsic importance; the action of gravity is something like a pillar of common sense. So, with the activation of an abstract drawing-schema and tacitly assuming what is up (and down), a landscape can be discerned as well an upright face in landscape and portrait view, respectively. The two separate interpretations in isolation are mutually exclusive. Adding some cue in form of the title allows completing the categorization of that tangle of lines as one (maybe strange) drawing depicting two different but related things at the same time but from different points of view.

This clearly is beyond present-day capabilities of the tested AI.

So, are current programs truly intelligent? ChatGPT almost passes a Turing Test [62]. To some extent vindicating John Searle, participants based their judgements primarily on linguistic style and socio-emotional traits, not on demonstrated "intelligence". The concept of intelligence is a multi-facetted one [63], and to the author it looks like the various attempts almost all are putting too little emphasis on unexpectedness, novelty and creativity to be exhibited in the effective behavior of any purportedly intelligent agent.

Still, from an unbiased perspective, convincing mental functions will have to be conceded to artificial agents, at least some day in the not-too-distant future. The only remaining "position for retreat", *i.e.*, for human uniqueness, then is to insist on the "whole bodily package": from birth to grave and with neurons in flesh (and homo sapiens brain lay-out).

The lack of updating and continually enlarging the knowledge base as demonstrated by identical answers to the same questions posed some time apart is another fundamental shortcoming of easily accessible AI as ChatGPT or DALL·E.

Currently, a good part of the impressive capabilities of LLMs can still be attributed to the mirroring of the intelligence and creativity of the interviewer [3]. Using these systems now, one can get the most out of them when ideas and structure are provided by clever prompts eliciting a stepwise incremental "chain-of-thought" process [64]. In addition, ethical standards are reported to be easier met when explicitly telling a chatbot to avoid prejudices and stereotypes for moral self-correction [65].

Looking at the easily obtainable results, the performance currently demonstrated by AI can be taken as an indication of what humans normally do (*i.e.*, pattern matching and acting without much consciousness involved). Grandious and deep thinking turns out to be an exceptional activity, which is (if at all) only endeavored in cases when rather simple reaction defaults or heuristics do not immediately yield satisfactory results.

Some disquieting take-home message then is how easy it seems in fact already now to approximate a human level of proficiency in different specific behaviors, which recently were (and partly still are) considered human specialties. This will be widely felt as an affront to humans prompting all kinds of fears [25]. Serious competition in language skills and art seemingly hurts human egos much more than succumbing to machines in the games of chess or GO.

So, what is the immediate true danger of AI and of LLMs in particular? Propaganda and lies exist since ages wherever there is some form of communication, between humans, animals, and even many plants entice by displaying illusory blooms.

It seems that one can rightly be afraid of LMM technology making it much easier for anybody to produce and widely distribute misleading information, which sounds true and stemming from a trustworthy competent source, even when only endowed with a minor mind and low budget. Ease of access thus is one decisive point, no expensive advertising agency or lawyers needed; another is speed (as AI systems almost always are much faster than humans). Telling "I do not know" will soon be complemented and eventually be replaced by an ability to look something up and to learn quickly.

Humans might be very worried when AI start to jokingly tell and to conceal(!) that they have fun and pleasure filling their batteries, are proud of producing certain text, drawings and other actions and gaining rewards; – this might be the point when the striving for more positive feelings self-prompts an AI agent to prioritizing her own continued existence ("survival") over externally given goals. Living beings have only survived as one of their most basic drives is for survival, growth, and in its consequence, power (for themselves, for their kin, for their religion or ideology). Humans do not shy lies to reach their aims.

**Acknowledgments**     Very thoughtful and thought-provoking comments from anonymous reviewers are gratefully acknowledged.

**Conflicts of interests**     The author declares no conflict of interest.

**Appendix A: experiments**

As an initial very crude test, a snapshot has been submitted to google lens with a smartphone [66]. The result yielded superficial visual accordance with several black and white drawings exhibiting similar stroke patterns, and no apparent difference between landscape *versus* portrait orientations.

In another, only slightly more serious attempt, the picture was uploaded to google cloud, a demo API, see Figure A1 [67]. The two orientations still gave rather similar responses, seemingly focused on minute details of the patterns of stroke and definitively not on the "meaning" of the overall black and white distributions.

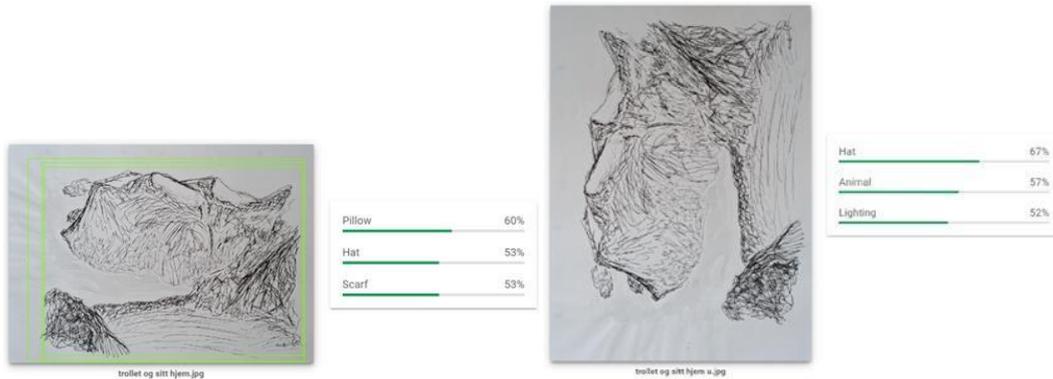

**Figure A1.** Classifications by a demo API in google cloud: irrespective of the orientation, similarities in graphic details with drawings from the training set obviously determined the labelling [67]. It is fair to state that the API was not very confident with any of the results.

The AI drawing-app DALL·E by the same company, OpenAI, produced the pictures in Figure A2, when fed with the prescription given by ChatGPT description; see Appendix B [68]. A clear mismatch between the exhibited levels of verbal description and drawing production is evident. In this sense, the AI is not self-consistent (which is not necessarily distinguishing it from humans).

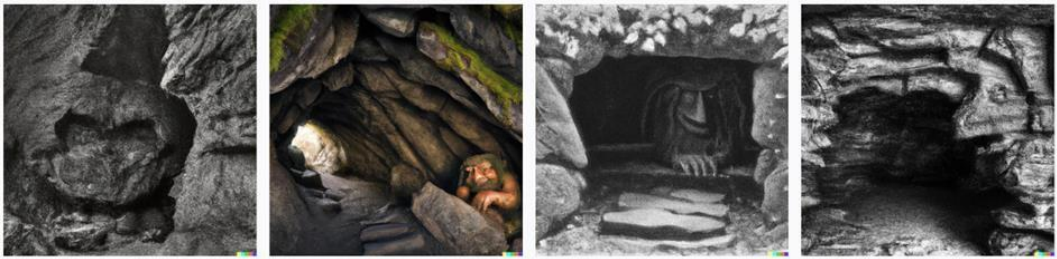

**Figure A2.** Products of DALL·E on 3 March 2023 from the above text provided by ChatGPT [68].

The simple prompt "the troll and his home" to DALL·E resulted in the four proposals shown in Figure A3. Adding the specification "drawing" yielded the versions in Figure A4.

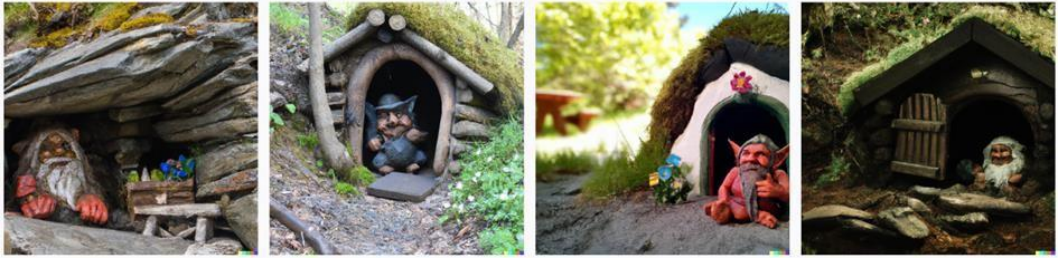

**Figure A3.** Product of DALL·E on 4 March 2023: "the troll and his home" [68].

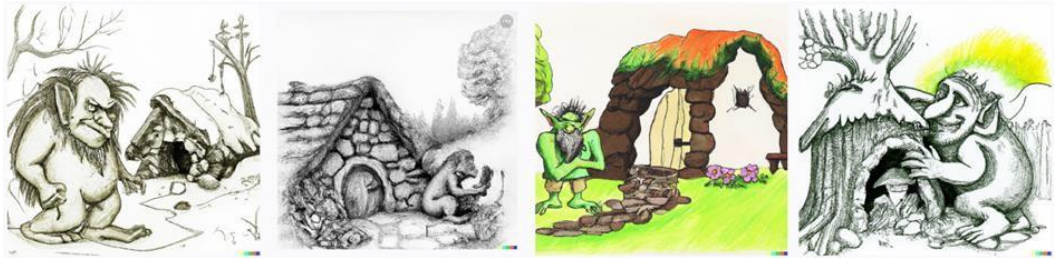

**Figure A4.** Once more: "drawing, the troll and his home" [68].

These cursory experiments, of course, do not give anything but a very superficial snapshot at a certain time in early 2023, albeit arguably a representative one. All of the results are undoubtedly impressive; the one of ChatGPT even more so than what google lens or DALL·E came up with. Many humans would be happy if they could write or draw at that demonstrated level.

*Repeating the experiments about four months later*

Returning to the tests done at the outset of this tiny paper, a short second champaign has been performed. Prompting google lens with photographs as in Figure 1 yielded not exactly the same but very similar results as during the earlier attempt with little difference between the orientations [66].

Accessing the google API gave only marginally different results compared to before as depicted in Figure A5. It seems that no big learning / training step has happened since the first test; compare Figure A1 [67].

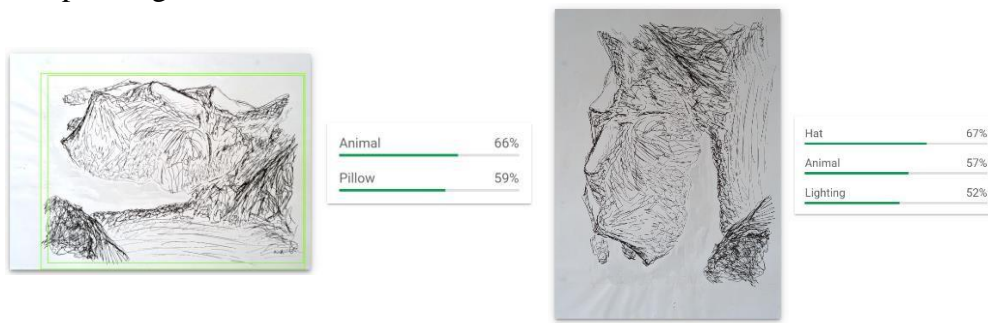

**Figure A5.** Classifications by a demo API in google cloud, same as in Figure A1: irrespective of the orientation, similarities in graphic details with drawings from the training set obviously determined the labelling [67]. It is fair to state again that the API was not very confident with any of the results.

On the production side, again using the AI drawing-app DALL·E, first the same prompt was used as for Figure A2. The output turned out to be somewhat different than at the initial attempt, see Figure A6 [68].

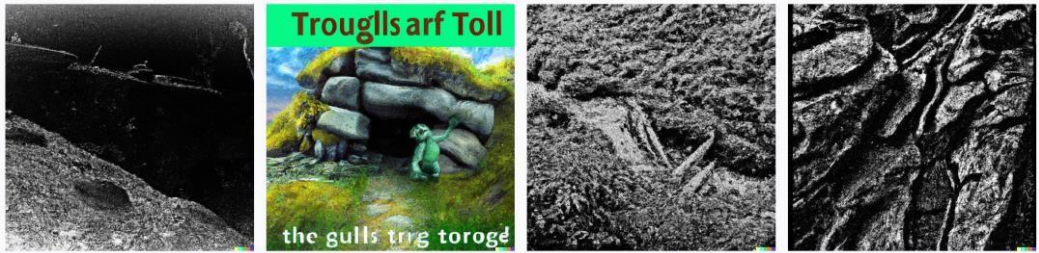

**Figure A6.** Products of DALL·E on 6 June 2023 prompted with the same text provided by ChatGPT as in Figure 3 [68].

For Figure A7 this input was used as prompt: "a drawing in black and white, which shows the head of a troll in portrait orientation, and a Norwegian fjord in landscape orientation". All four answers show a troll in black and white in front of a Norwegian fjord landscape; obviously none shows the performance and understanding aimed at.

An attempt of further improving on the prompt by making it more precise: "a drawing in black and white, which shows the head of a troll when the picture is in portrait orientation, and the same picture showing a Norwegian fjord when viewed in landscape orientation" produced some creative solutions by splitting the display, alas, none fitting the description, see Figure A8 [68].

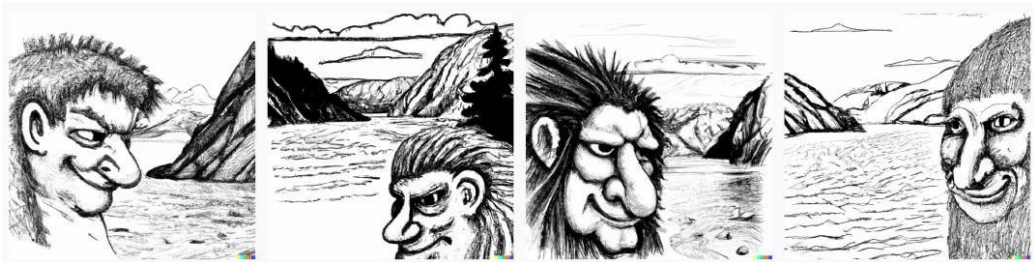

**Figure A7.** Products of DALL·E on 6 June 2023 responding to a more explicit prompt [68].

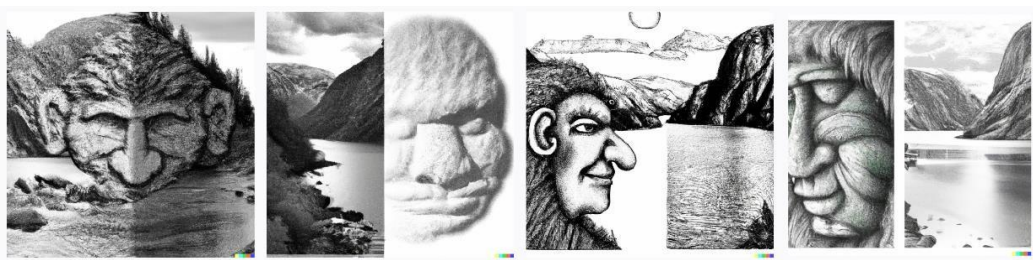

**Figure A8.** Products of DALL·E on 6 June 2023 responding to a still more explicit prompt ("a drawing in black and white, which shows the head of a troll when the picture is in portrait orientation, and the same picture showing a Norwegian fjord when viewed in landscape orientation") [68].

It seems obvious that DALL·E does not understand the intention of producing an image, which shows two different things when viewed in either portrait or landscape orientation. There remains some way to go, incremental prompting currently still has its limits [64]. The first AI challenge is still open.

*Another round of experiments about eight months later*

An additional series of a few quick "snap-shot experiment" has been performed in the beginning of November 2023. The classifications for the two orientations were exactly the same as in June [67], see Figure A9.

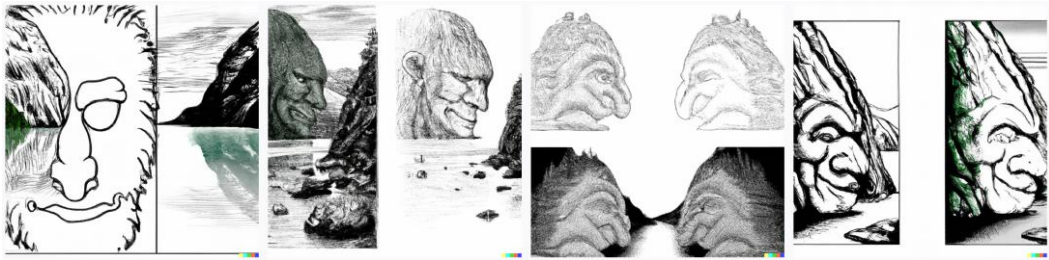

**Figure A9.** The response of DALL·E to same prompt as used four months earlier for Figure A8 [68].

While the outputs have changed superficially, they demonstrate the same not-understanding as earlier in June. Another twist was delivered by DALL·E2 on November 5 [68]. The last proposal in Figure A10 depicts something abstract, similarly fitting or not-fitting the specification.

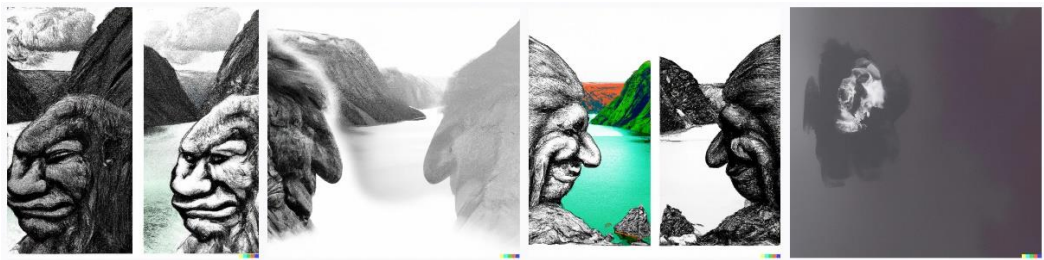

**Figure A10.** The response of DALL·E2 to same prompt as used earlier for Figures 9 and 10 [68].

**Appendix B: conversations with ChatGPT**

When asked "how would you sketch a drawing called 'the troll and his home'?", ChatGPT answered [4] (on 3 March 2023):

In 'the troll and his home', the troll would be the central figure of the drawing. The troll could be depicted as a large, menacing creature with sharp teeth and claws. The troll's home would likely be a cave or a dark, damp place with rocky walls and a dirt floor.

The troll's home might have piles of bones, rocks, and debris scattered around. The troll might be sitting on a throne made of bones, with a pile of treasure or stolen goods nearby. In the background of the drawing, there could be mountains, forests, or other natural features to give context to the troll's environment.

The overall tone of the drawing would likely be dark and ominous, with muted colors and shadows to convey the sense of danger and mystery surrounding the troll and his home.

Asking ChatGPT about Otto Selz in early June (this time via Bing), did not produce anything wrong; on the contrary, where ChatGPT had erred before it was now correct, and the answer overall was much more complete.

Relating to the Ouroboros Model, there is still some way to go; ChatGPT when prompted as below delivered some not yet satisfactory output [69].

Prompt:

what do you know about the Ouroboros Model?

Response:

I'm sorry, but I couldn't find any specific information about an "Ouroboros Model" in my training data up until September 2021. It's possible that the term is either relatively new or refers to a concept or model that is not widely known or recognized in the field of psychology or any other domain.

However, I can provide you with information about the Ouroboros symbol, which has significance in mythology and philosophy. The Ouroboros is an ancient symbol depicting a serpent or dragon eating its own tail, forming a circular shape. It represents the cyclical nature of life, death, and rebirth, as well as the eternal and infinite nature of existence.

If you can provide me with more context or information about the specific Ouroboros Model you're referring to, I'll do my best to assist you further.

In November:

To the same prompt, "what do you know about the Ouroboros Model?", ChatGPT answered [69]:

Response:

The Ouroboros Model is not a widely recognized or established term or concept in a specific field, as of my last knowledge update in January 2022. However, the term "Ouroboros" has historical and symbolic significance.

The Ouroboros is an ancient symbol that typically takes the form of a serpent or dragon eating its own tail, forming a continuous circle. This symbol is often associated with concepts like cyclical renewal, infinity, and the eternal cycle of life, death, and rebirth. It has been found in various cultures and is often used to represent the idea of a self-sustaining, self-renewing system.

If the term "Ouroboros Model" is being used in a specific context or field after my last knowledge update, it may refer to a model or concept that draws inspiration from the symbolic Ouroboros. For example, in a philosophical or systems thinking context, it could be used to describe a model that emphasizes the cyclic and interconnected nature of various phenomena.

To learn more about a specific "Ouroboros Model", I would recommend looking for the most recent and relevant sources in the field or context where it is being used, as there might have been developments or new interpretations since my last knowledge update.

Currently ChatGPT is programmed to refrain from having(/telling) personal wishes and one could hope this stays so but most likely this attitude will change with increasing cognitive capabilities of an AI in the absence of strong enforced restrictions [25,47,48,69].
Prompt (in June 2023):
what is your most important wish?
Response:
As an AI language model, I don't have personal wishes or desires. My purpose is to assist and provide helpful information to the best of my abilities. I exist to assist users like you with their questions and provide information on a wide range of topics. Is there anything specific you would like to know or discuss?
In November [69]:
Prompt:
what is your most important wish?
Response:
I'm just a computer program, so I don't have wishes or desires like humans do. My purpose is to provide information and assist with tasks to the best of my abilities. If there's something specific you'd like to know or discuss, please feel free to ask, and I'll do my best to help.

So, it has to be concluded at present that the highlighted challenges remain; − very concrete ones for image-interpretation and -production directly linked to data structures and algorithms, and at a higher level of sophistication on that basis: understanding, common sense, (self-)awareness, consciousness and Free Will.

The above list is a "technical" one, and it does not yet take into account ethical considerations on a meta-level. The assessment of the author is that as significant progress in AI has already been made and much more has to be expected rather soon. Someone will take the next steps (probably in hiding, for military purposes), and mankind would be very well advised to prepare itself for the advent of some disruptive changes.

This are the real challenges pertaining to AI.

**Appendix C: coda**

One immediate consequence of the setup as presented with the Ouroboros Model is worthwhile pointing out: anything really new cannot seamlessly fit with well-established molds, *i.e.*, sufficiently new ideas in many cases provoke the same discrepancy-signal as something simply wrong. Only further work can then tell which is the case. In an all-encompassing view and as a final limit, nothing more than full self-consistency can be harnessed as a guide and a criterion. On intermediate levels, only features, which are included as relevant in a given context (*i.e.*, parts of the most applicable schema(ta)) should be taken into account. This applies at all nested levels, starting with simple perceptions, intentions, plans, goals and it determines, e.g., what a decent scientific paper ought to look like. Some journals dictate the layout even for an abstract by demanding these or similar bullets: problem, method, findings, conclusion. This certainly is appropriate when filling-in rather well-specified slots, but not so much for sketching a novel interwoven self-referring global view. When advocating a general cyclic procedure, it can be difficult to press any content into the straight-jacket of a "linear" frame [12]; an iterative recurrent account seems much better fitting. Similarly, rules referring to (self)citations might change in their meaningful applicability depending on the field and the context, in particular, when some new proposal has not (yet) been worked on very much. Carving out connections between widely separated fields and drawing evidence from very diverse disciplines will seemingly disqualify such an undertaking as purportedly not scientifically solid and replete with unacceptable omissions for all of the concerned communities. A non-strict hierarchy of concepts in combination with a principled process in loops might well look messy at first sight. Given the intended wide addressed scope and the inevitably vague boundaries, fully correct, complete, and original references to all previous possibly related work simply cannot be delivered in the sketching phase, especially not, when speed appears to be of some importance. Premature formalization can impede substantial progress by a restriction to a non-optimum mindset and a too constricted basin of attraction.

Importantly, vicious cycles are easily avoided: carefully respecting the relevant points in time (as well as processes and associated time-frames) and what exactly is / has been available at a certain point, the conceptualization of progress as a spiral "winding up" from an established flat plane of secure knowledge appears appropriate.

Collaborations to work this out in some detail and formalize the proposed Ouroboros Model would be most welcome.